\documentclass{article}
\usepackage{ijcai24}

\usepackage{times}
\usepackage{soul}
\usepackage{url}
\usepackage[hidelinks]{hyperref}
\usepackage[utf8]{inputenc}
\usepackage[small]{caption}
\usepackage{graphicx}
\usepackage{amsmath}
\usepackage{amsthm}
\usepackage{booktabs}
\usepackage{algorithm}
\usepackage{algorithmic}
\usepackage[switch]{lineno}
\usepackage{multirow} 
\usepackage{xcolor}  
\usepackage{makecell}
\usepackage{framed}
\usepackage{listings}
\definecolor{shadecolor}{gray}{0.9}
\usepackage{tcolorbox} 
\linenumbers

\newcommand{\guangran}[1]{{\color{black} #1}}
\newcommand{\wenzhe}[1]{{\color{black} #1}}

\title{Empowering Large Language Models on Robotic Manipulation \\ with Affordance Prompting}

\author{
Guangran Cheng*$^1$
\and
Chuheng Zhang*$^2$\and
Wenzhe Cai$^{1}$\and
Li Zhao$^2$\and
Changyin Sun$^1$\And
Jiang Bian$^2$\\
\affiliations
$^1$Southeast University\\
$^2$Microsoft Research Asia\\
\emails
chenggr@seu.edu.cn
}

\begin{document}
\nolinenumbers
\maketitle

\begin{abstract}
While large language models (LLMs) are successful in completing various language processing tasks, they easily fail to interact with the physical world by generating control sequences properly.
We find that the main reason is that LLMs are not grounded in the physical world.
Existing LLM-based approaches circumvent this problem by relying on additional pre-defined skills or pre-trained sub-policies, making it hard to adapt to new tasks.
In contrast, we aim to address this problem and explore the possibility to prompt pre-trained LLMs to accomplish a series of robotic manipulation tasks in a training-free paradigm.
Accordingly, we propose a framework called LLM+A(ffordance) where the LLM serves as both the sub-task planner (that generates high-level plans) and the motion controller (that generates low-level control sequences).
To ground these plans and control sequences on the physical world, we develop the \textit{affordance prompting} technique that stimulates the LLM to 1) predict the consequences of generated plans and 2) generate affordance values for relevant objects.
\guangran{Empirically, we evaluate the effectiveness of LLM+A in various language-conditioned robotic manipulation tasks, which show that our approach substantially improves performance by enhancing the feasibility of generated plans and control and can easily generalize to different environments.}
\end{abstract}

\section{Introduction}
Recent large language models (LLMs)~\cite{ouyang2022training,chowdhery2022palm,chung2022scaling} have exhibited remarkable capabilities in a wide range of natural language processing (NLP) tasks.
As these pre-trained models assimilate extensive knowledge from internet-scale text corpora and various domain-specific datasets, they can be leveraged as foundation models and provide rich prior knowledge. 
Owing to these advancements, new paradigms have emerged, employing LLMs as artificial intelligence (AI) assistants to perform embodied robotic tasks in real-world settings~\cite{driess2023palm,brohan2023rt,vemprala2023chatgpt}. 

We study the problem of language-conditioned robotic manipulation control with LLMs.
Existing approaches can be generally classified into two main categories: employing LLMs as high-level sub-task planners~\cite{ahn2022can,huang2022language,wang2023describe}, or directly training an end-to-end large model as low-level motion controllers~\cite{driess2023palm,brohan2023rt,mu2023embodiedgpt}.
The first category leverages the powerful planning and reasoning capabilities of LLMs, enabling them to decompose human instructions into a series of textual sub-tasks. 
Nevertheless, to execute these decomposed plans in real-world scenarios, current methods still depend on pre-trained skills or primitive actions~\cite{ahn2022can,huang2023voxposer,singh2023progprompt}, which are usually learned by behavior cloning or reinforcement learning.
This reliance on specific sub-policies can limit the applicability of these approaches, as they necessitate vast amounts of robotic data and often struggle to generalize to unseen environments~\cite{huang2023voxposer} and different embodiments~\cite{Bousmalis2023RoboCatAS}.
The second category typically trains a large-scale multi-task backbone model that integrates both linguistic and visual modalities to generate end-to-end control sequences~\cite{brohan2023rt,driess2023palm,reed2022generalist}.
However, the development of such models requires extensive multi-modal datasets, encompassing a diverse array of robotic tasks. 
This is costly for many scenarios.
Therefore, considering the impressive commonsense knowledge and powerful in-context learning abilities demonstrated by LLMs, can LLMs function as both sub-task planners and motion controllers, thereby addressing robotic manipulation tasks without the need for additional training?

\begin{figure*}[t]
	\centering
	\includegraphics[width=0.8\linewidth]{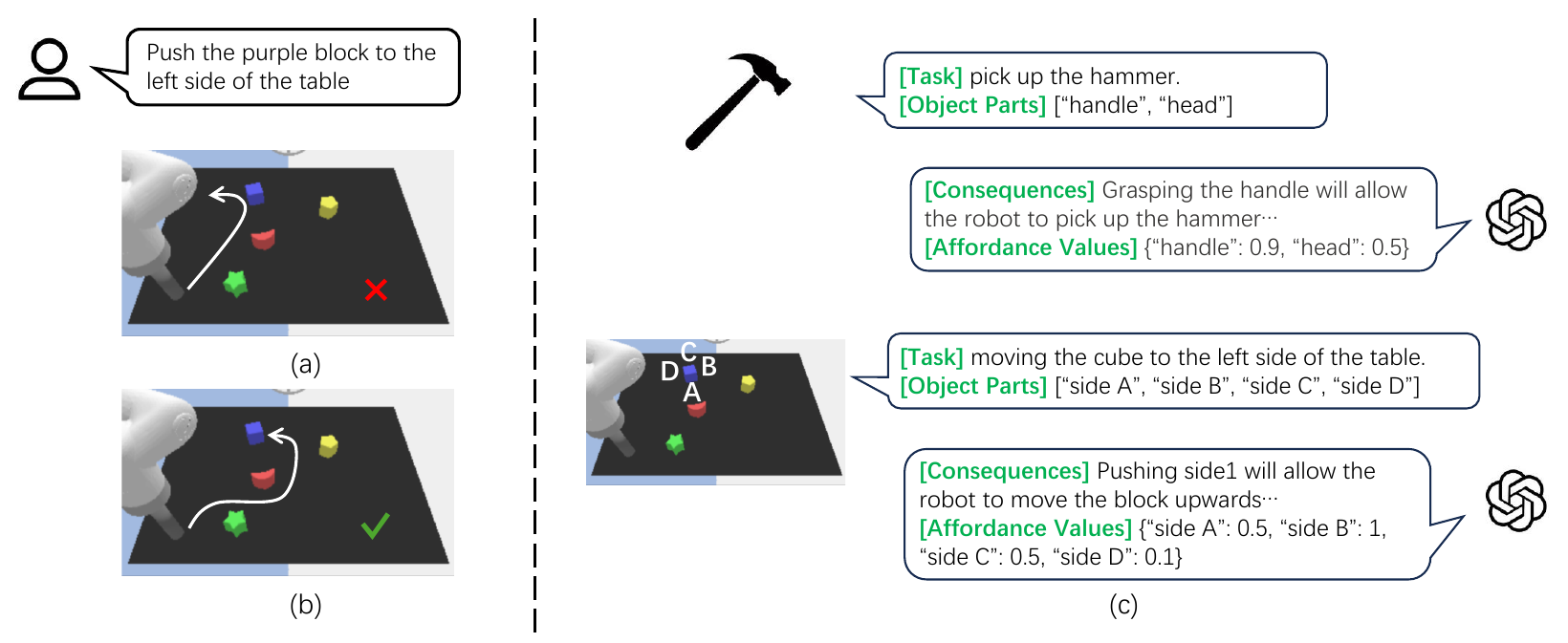}\\
	\caption{Consider the task of ``Push the purple block to the left side of the table''. When the control sequences generated from LLMs are not grounded in the physical world, the robot will move to the left side of the block to push it to the left (a) instead of the right location (b). This is due to the gap between the physical world and generated language plans. This gap can be bridged by prompting LLMs to predict execution consequences and goal-conditioned affordance values (c) in the proposed LLM+A method.}\label{fig1}
\end{figure*}

However, how to exploit commonsense from LLMs to real-time fine-grained robotic manipulation tasks remains a challenging problem. 
This difficulty primarily comes from the fact that LLMs are not grounded in the physical world, which can potentially result in erroneous or inexecutable plans~\cite{ahn2022can,yao2022react}.
For instance, consider a tabletop robotic arm situated to the left side of the table, and the instruction is ``push the purple block to the left side of the table''.
The arm may move directly right and then push the block as shown in Figure \ref{fig1}(a) instead of maneuvering around to the right side of the object as shown in Figure \ref{fig1}(b).
In this scenario, the block will not move as expected.
The main reason is that LLMs lack more comprehensive information regarding the current environment, such as the spatial relationship on the objects. 
Besides, pre-trained LLMs neglect to comprehend the consequences of the generated plan in the actual physical world.
Therefore, to generate executable control sequences, LLMs must consider physical laws and object functionalities.
This can help robots predict the consequences of generated actions when interacting with objects or specific environmental features.

In the field of robotics, the concept of \textit{affordance} is considered a crucial mechanism to enable robots to understand and interact with environments. 
Given the task instruction, affordance values indicate the functional priorities for the robot of objects in the current environment to complete the task.
At the current stage of affordance research, the related prior knowledge is usually provided by humans~\cite{yang2023recent}.
In contrast to previous studies, we demonstrate that LLMs are proficient at predicting execution consequences and inferring affordance values of different object parts as shown in Figure \ref{fig1}(c), which effectively provides robots with useful information to complete instructions.
For instance, within the context of hammering or pushing tasks, LLMs can interpret human directives and ascertain task-specific executable parts for grasping, functioning, effecting, or pushing respectively.

Motivated by this concept, we introduce a framework, called LLM+A, to exploit the extraction of embodied commonsense and reasoning capabilities of LLMs to generate control sequences for robotic manipulation tasks following textual instructions in a training-free pradigm.
Firstly, we employ pre-trained vision language models (VLMs), such as open-vocabulary detectors~\cite{minderer2022simple,liu2023grounding}, or large multimodal models ~\cite{gao2023llama,awadalla2023openflamingo}, to provide textual observation of target objects (such as object shapes, colors, and positional relationships) and interactive environments to the LLMs.
Then, we obtain goal-conditioned affordance values from LLMs that describe the priorities of object executable parts for interaction via \textit{affordance prompting}.
Based on the above visual perception and affordance values, LLMs decompose human instructions into high-level sub-tasks, which are feasible in the physical world.
Subsequently, LLMs also generate control sequences for the current task, such as waypoints of the robot end-effector.
Our experiments demonstrate that grounding LLMs in the physical world to generate motion plans via unlocking their affordance knowledge can highly enhance the performance in robotic manipulation tasks compared to non-grounded baselines.

Our contributions are summarized as follows:
\begin{itemize}
    \item[$\bullet$] We propose LLM+A that adopt large language models (LLMs) to serve as both the high-level sub-task planner and the low-level motion controller in robotic control tasks in a training-free paradigm.
    \item[$\bullet$] To improve the physical executability of both the sub-task plans and the control sequences generated while adhering to the language instruction, we propose \textit{affordance prompting} to stimulate the ability of LLMs to infer goal-conditioned affordance values, which indicate the executable priorities of different parts of interacted objects.
    \item[$\bullet$] Experimental results on heterogeneous robotic tasks validate the effectiveness and robustness of our method.
\end{itemize}

\section{Related work}
\textbf{LLMs for Sub-Task Planning and Motion Controlling.}
With the development of transformers in recent years, pre-trained large language models~\cite{ouyang2022training,chowdhery2022palm,chung2022scaling} have become an active area of research.
These models, pre-trained on vast amounts of internet-scale text corpora from various tasks, exhibit remarkable commonsense and reasoning capabilities in embodied tasks. 
Numerous recent approaches successfully employ LLMs to decompose abstract and human instructions into natural language-based high-level plans~\cite{ahn2022can,huang2022language,ding2023task,ren2023robots} or code-based plans~\cite{liang2023code,huang2023instruct2act}.
For instance, ZSP~\cite{huang2022language} demonstrate that LLMs can be utilized for task planning in household domains through iteratively augmented prompts, enabling the semantic translation of plans into admissible skills.
Similarly, SayCan~\cite{ahn2022can} leverage LLMs to facilitate robot task planning by incorporating affordance functions to ensure plan feasibility.
While these methods show surprising zero-shot generalization ability of task planning, the execution of decomposed plans remains dependent on pre-trained skills, which are usually acquired via behavior cloning and reinforcement learning.
This reliance may limit their applicability, as the training process necessitates substantial amounts of robotic data, which is costly to obtain.
On the other hand, some general robotic models have been proposed to achieve end-to-end control for real-world robotic applications~\cite{brohan2023rt,brohan2022rt,driess2023palm,mu2023embodiedgpt,stone2023open}.
These methods benefit from high-capacity networks and open-ended task-agnostic training with diverse datasets.
In contrast, the proposed LLM+A leverages LLMs as both the sub-task planner and the motion controller in a training-free paradigm in robotic manipulation tasks.

\textbf{Affordance for Robotics.}
As a popular concept proposed in the field of psychology, affordance has been extensively utilized in robotic tasks to facilitate agents' comprehension and interaction with dynamic environments~\cite{xu2021deep,wu2021vat,mo2021where2act}.
Briefly, affordance encapsulates the potential outcomes and effects resulting from robot's actions on a specific object or, more broadly, a segment of the environment. 
Existing research can be divided into three primary categories: modeling action possibilities~\cite{mo2021where2act,borja2022affordance}, generating keypoint affordances~\cite{qin2020keto,xu2021affordance}, and learning partial dynamic models exclusively for afforded actions~\cite{xu2021deep,khetarpal2021temporally}. 
Since their development primarily relies on data-driven approaches within the visual domain, these methods exhibit limitations when applied to language-conditioned scenarios.
Recent methods begin to leverage LLMs to obtain language-conditioned affordance values.
For example, HULC++~\cite{mees2023grounding} employed a self-supervised visual-lingual affordance model to guide robots toward actionable areas referenced by language.
VoxPoser~\cite{huang2023voxposer} extracted affordances and constraints for robotic manipulation tasks from pre-trained LLMs, demonstrating generalizability to open-set instructions.
In this work, we concentrate on grounding LLMs in more fine-grained robotic tasks by predicting execution consequences and extracting physical affordance values.

\section{Method}
In this section, we first introduce the formulation of robotic manipulation tasks (Sec. \ref{Problem formulation}). 
Then, we describe the LLM+A framework where the pre-trained visual-language model (VLM) serves as the observation descriptor and the large language model (LLM) serves as the high-level sub-task planner and the low-level motion controller (Sec. \ref{Task and motion plan from LLMs}).
Later, we introduce \textit{affordance prompting} to predict consequences and generate affordance values to bridge the gap between generated plans/control sequences and the physical world (Sec. \ref{Affordance from LLMs}).

\subsection{Robotic Manipulation}\label{Problem formulation}

In robotic manipulation tasks that we consider in this paper, the LLM-based agent needs to generate control sequences for a tabletop robotic arm to complete a given task instructed in open-vocabulary natural language based on the image observation.
Specifically, on the $t$-th time step, the agent $\pi(a_t|o_t, l)$ perceives the image observation $o_t$ and outputs the action $a_t$ to follow the instruction $l$.
The instruction $l$ is not constrained by any templates, grammatical structures, or vocabularies.
For example, the instruction can be ``push the red block to the left center side of the table'' or ``separate the yellow block and the green block''.
In this paper, the action of the LLM-based agent is a planned path segment of the end-effector of the robotic arm, represented by $K$ coordinates specifying the waypoints of the end-effector, i.e., when $a_t = \left( (x_1, y_1), \cdots, (x_K, y_K) \right)$, the end-effector will move along the path $(x_1, y_1) \to \cdots \to (x_K, y_K)$.

\iffalse
Considering robotic manipulation tasks given human textual instruction $l$ and current image observation $s_t\in\mathcal{S}$, we first query VLMs with $s_t$ to obtain relative textual information $o_t$. 
Our goal is to utilize LLMs to generate executable control policies $\pi(\bold{\tau}|o, l)$ for robotic manipulation tasks. 
In particular, we focus on developing open-vocabulary language-conditioned visuomotor policies, which means the instruction $l$ is not constrained by any predefined templates, grammatical structures, or vocabularies.
As a result, textual instructions can vary, such as ``push the red block to the left center side of the table'' or ``separate the yellow block and the green block''.
Instead of relying on any pre-trained policies or primitive actions, we utilize LLMs to generate the textual robot trajectories at each time step $t$, i.e., $\pi_t$.
Importantly, the motion plan is updated online, in accordance with changes in the observation.
\fi

\subsection{LLM+A}\label{Task and motion plan from LLMs}
\begin{figure*}[t]
	\centering
	\includegraphics[width=0.8\linewidth]{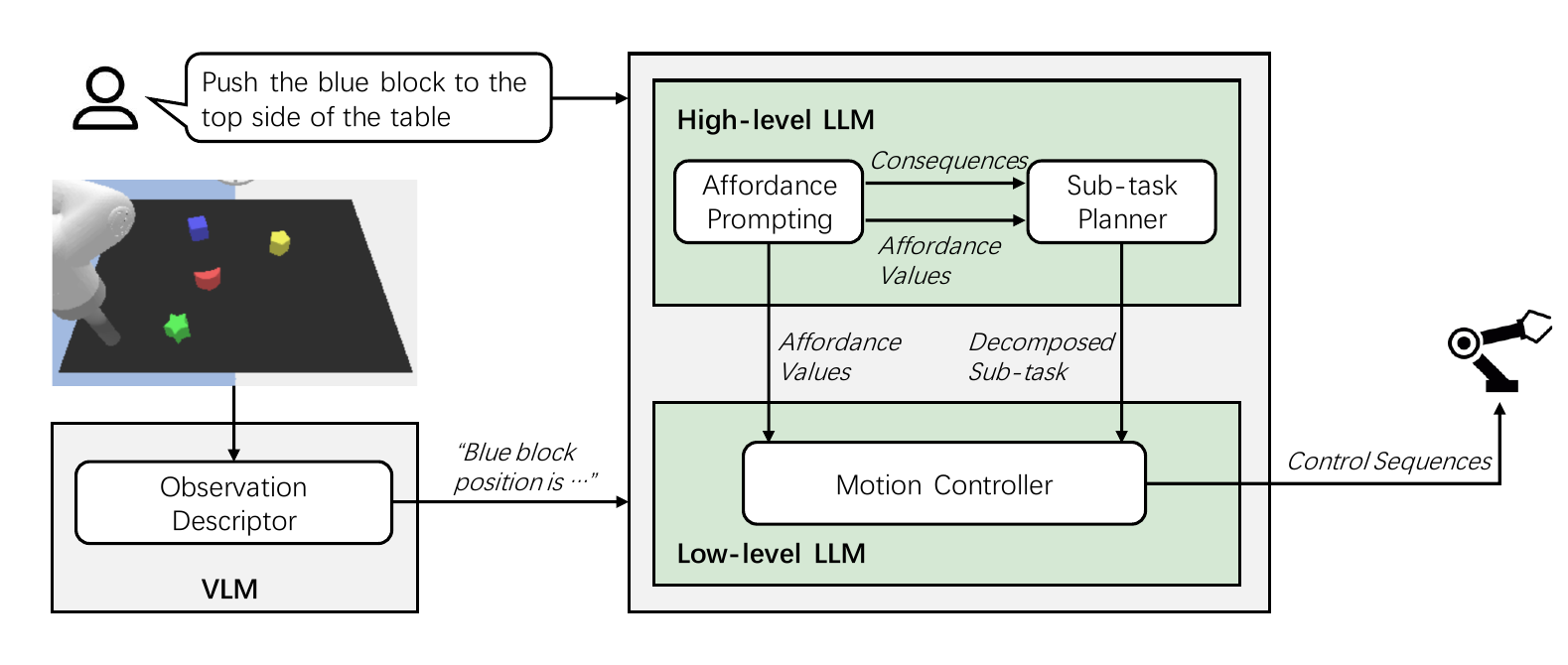}\\
	\caption{Overview of LLM+A. Given language instructions and image observations, LLM+A produces sub-task plans and control sequences for robotic control tasks. LLM+A is composed of a VLM and a hierarchical LLM. The VLM serves as an observation descriptor to provide the environment perception to the LLM. The high-level LLM is responsible for sub-task planning and the low-level LLM for motion controlling. Notably, the affordance values from the high-level LLM are necessary intermediate information for the LLM to understand the effects of potential actions and generate feasible plans grounded in the physical world.}
    \label{framework}
\end{figure*}

In the LLM+A framework, we aim to leverage the commonsense knowledge and reasoning/planning capability from both the pre-trained VLM and LLM to complete robotic manipulation tasks.
Further, we develop \textit{affordance prompting} to incorporate the concept of affordance into zero-shot prompting for the LLM, which can be regarded as an extension of chain-of-thought (CoT) to embodied robotics.
We present the LLM+A framework in Figure \ref{framework} and introduce different modules of LLM+A as follows.

\textbf{Observation Descriptor.}
In this module, we feed the current image observation $o_t$ combined as a designed prompt template to the VLM-based observation descriptor to generate text description $s_t$ to provide necessary information needed in the subsequent decision making process of the LLM-based sub-task planner and motion controller.
Specifically, the text description $s_t$ contains the spatial location of relevant objects and all functional parts of them such as the four sides of the block.

\textbf{Sub-task Planner.} 
In this module, we feed the text description of the current observation $s_t$ as well as the instruction $l$ to the sub-task planner to obtain a series of planned high-level sub-tasks $g=(g_1, g_2, \cdots)$ required to accomplish the task based on the current observation.
Notice that this plan can change on the subsequent time steps based on future observations and the agent typically only executes the first sub-task on the current step.
Further, to incentivize the sub-task planner to generate feasible plans, we develop the \textit{affordance prompting} technique that instructs the sub-task planner to predict the expected consequences of the control and generate affordance values for the functional parts of relevant objects.
%In this paper, we use GPT-4~\cite{openai2023gpt4} for the sub-task planner.
We present the simplified version of the prompt template as follows:

\begin{tcolorbox}[colback=gray!15, % 灰底颜色，值越小颜色越深，范围在0到100之间  
                 arc=3mm, % 框的圆角半径  
                 boxrule=0pt, % 框的边框线宽，设为0表示不显示边框  
                 left=2mm, right=2mm, top=2mm, bottom=2mm, % 设置内边距  
                 boxsep=2pt % 设置文本和框之间的间距  
                ]  
Template for Sub-Task Planner: \textit{\\
You are a robotic arm on the table which can [arm skills].\\
You need to accomplish a series of robotic manipulation tasks: [guidelines]. \\
The task instruction is [task instruction].\\
The objects on the table are [object parts].\\
You need to: \textcolor{purple}{\\
1. output the \textbf{consequences} of potential actions; \\
2. output the \textbf{affordance values} of each object parts considering the potential consequences;}\\
3. output the \textbf{decomposed sub-tasks} according to the consequences and affordance values.
} 
\end{tcolorbox}  

In the template, 
\textit{[arm skills]} describe the functions of the type of the arm; 
\textit{[guidelines]} describe guidelines or contexts of the task such as the orientation of the table, the range/orientation of the spatial coordinates;
\textit{[task instruction]} is the open-vocabulary natural-language-based instruction $l$;
\textit{[object parts]} are text description of the observation $s_t$ generated by the observation descriptor.
At last, we ask the LLM to decompose the task instruction $l$ into sub-tasks $g$.
\textit{Decomposed sub-tasks} refer to more specific plans appropriate for the robot to execute starting from the current state, e.g., ``approach right side of the green block while avoiding the red block'' and ``push the green block to the top''.
To generate more feasible sub-tasks, we adopt \textit{affordance prompting} which is highlighted in red.
\textit{Consequences} refer to the effects of object parts after the physical interaction by possible skills of the robot and \textit{affordance values} indicate the extent to which each part of the object is expected to be interacted.
These concepts will be further explained later in Sec. \ref{Affordance from LLMs}.
% Note that in this process, the affordance prompting technique is zero-shot in the sense that we do not provide any examples in the prompt to regulate the form of the output.
% This zero-shot usage makes the technique easily extend to different tasks.

\iffalse
Where [Guidelines] describes certain
guidelines and contexts of the current scene, such as the orientation of the table and the coordinates of the boundaries.
[Object Parts] are visual captions from the VLM and [Possible Skills] depend on the specific types of the arm.
The highlighted [Consequences] indicate the effects of object parts after physical interaction with possible skills and [Affordance Values] denote the affordance value for each side of the relevant objects, which will be further explained in Sec. \ref{Affordance from LLMs}. 
[Decomposed Tasks] refer to more specific sub-tasks appropriate for robotic execution, such as ``approach right side of the green block while avoiding the red block'', and ``push the green block to the top''.
Notably, in the sub-task planning process, we do not include any examples in the prompt to provide additional information.
This zero-shot setting makes our method easily generalize to different tasks.
\fi

\textbf{Motion Controller.}
In this module, given the decomposed sub-tasks $g=(g_1, g_2, \cdots)$ and the affordance values associated with different object parts, we ask the LLM to generate the specific action $a_t$. 
We also present the simplified version of the prompt template of this procedure as follows:

\iffalse
Finally, given $[l_1, l_2, ...]$ and $[a_1,a_2,...]$, we query the LLM to generate the specific motion control sequences $\pi_t=[\tau_0,\tau_1,...,\tau_T]$ for the current time step.
This is also achieved by employing a prompt, similar to the following template:
\fi

\begin{tcolorbox}[colback=gray!15, % 灰底颜色，值越小颜色越深，范围在0到100之间  
                 arc=3mm, % 框的圆角半径  
                 boxrule=0pt, % 框的边框线宽，设为0表示不显示边框  
                 left=2mm, right=2mm, top=2mm, bottom=2mm, % 设置内边距  
                 boxsep=2pt % 设置文本和框之间的间距  
                ]  
Template for Montion Controller: \\
\textit{You are a robotic arm on the tabletop which can [arm skills]. \\
You need to accomplish a series of robotic manipulation tasks: [guidelines]. \\
The task instruction is [task instruction].\\
The objects on the table are [object parts].\\
Given the [decomposed sub-tasks] and the \textcolor{purple}{[affordance values]}, you need to output the \textbf{control sequence}. \\
Note that [notes]. \\ 
The examples are as follows: [examples]. }
\end{tcolorbox} 

In this template, \textit{[decompoased sub-tasks]} and \textit{[affordance values]} are the outputs generated by the sub-task planner.
The control sequence is the action of the LLM-based agent which refers to a series of the waypoint coordinates of the end-effector in our case.
In addition, we also provide \textit{[notes]} and \textit{[examples]} to facilitate the LLM to generate better control sequences.
\textit{[Notes]} can include formatting instructions such as ``You need to generate the above outputs with JSON format'', 
and \textit{[examples]} follows the few-shot prompting practice.

\subsection{Affordance Prompting}\label{Affordance from LLMs}
In robotic tasks, the concept of affordance plays a crucial role in enabling robots to comprehend and interact with the corresponding physical environment.
This generally depends on prior knowledge of relevant actions and the task instruction.
% intentions that must be fulfilled.
LLMs are highly proficient at inferring affordance values, owing to their rich 
commonsense knowledge learned from diverse robotic-related datasets during pre-training.
For example, as shown in Figure \ref{fig1}(c), LLMs can accurately assess different parts of the hammer and their affordance values for the grasping task, based on which robot motions can be effectively suggested.
Another example is a robotic manipulation task, where LLMs identify the most actionable edge with the highest affordance value for pushing the block to the left side of the table. 
% Incited by this, 
Motivated by this observation, the goal of our \textit{affordance prompting} technique
is to unlock LLMs with the ability to generate goal-conditioned affordance value, serving as a feasible intermediate reasoning step that assists the robot in understanding action priorities.

In LLM+A, given \textit{[arm skills]}, \textit{[task instruction]}, and \textit{[object parts]}, we firstly query the LLM to generate \textit{[consequences]}, which reason about the future effects of possible actions.
Then, we ask the LLM to generate \textit{[affordance values]} of different object functional parts which indicate their usefulness in completing the given instruction. 

Note that in this process, the \textit{affordance prompting} technique is zero-shot in the sense that we do not provide any examples in the prompt to regulate the form of the output.
This zero-shot usage makes the technique easily extend to different tasks.
Empirically, the generated affordance values can effectively improve the feasibility of decomposed sub-tasks and control sequences from LLMs in the physical world.

To sum up, the \textit{affordance prompting} technique provides the following advantages to facilitate reasoning in robotic tasks:
\begin{enumerate}
\item First, the \textit{affordance prompting} technique assists LLMs in constraining control sequence updates within the set of feasible actions to follow the task instruction.
\item Secondly, the affordance values are convenient to obtain since only a single intermediate reasoning step is necessitated for LLMs, without requiring additional training or a fine-tuning process.
\item Thirdly, the \textit{affordance prompting} technique adopts a zero-shot setting, which is robust across various envrionments and capable of extending to heterogeneous tasks.
\end{enumerate}

\section{Experiments}
We conduct experiments in various robotic tasks to answer the following questions: 
1)~How effective is LLM+A for physical interaction compared to other state-of-the-art baselines?
2)~How well does LLM+A predict affordance values and plan action sequences?
3)~How robust is \textit{affordance prompting} when generalized to heterogeneous tasks? 

\subsection{Experimental settings}
\textbf{Implementation details.} 
For visual perception, given an image observation, we use the open-vocabulary detector Grounding DINO~\cite{liu2023grounding} to detect the bounding boxes of relevant objects, and use GPT-4 (June) from OpenAI API~\cite{openai2023gpt4} for both sub-task planner and motion controller. For the high-level sub-task planner, including the affordance prediction and consequence prediction, we do not provide any example outputs (zero-shot). For the low-level motion controller, we provide two examples in prompts to formalize the output style.
The detailed prompts are listed in the Appendix B.1. Instead of using our LLM+A to re-plan new trajectory waypoints every time step, the plan will be updated after the robots finish the $K=5$ waypoints in the previous round to increase the time efficiency.

\textbf{Tasks.}
\guangran{We evaluate LLM+A using eight simulated task tasks from Language-Table~\cite{lynch2023interactive} and CLIPORT~\cite{shridhar2022cliport}, covering three different manipulation skills (push, pick, and place):
1)~Block-to-Position (B2P): Push a block to an absolute location on the board (e.g., the top-left corner). 
2)~Block-to-Block (B2B): Push a block to another block. 
3)~Separate (Sep): Separate two blocks. 
4)~Stack-Block-Pyramid: Build a pyramid of colored blocks through the step-by-step language instructions.
5)~Towers-of-Hanoi: Solve the three-ring Towers of Hanoi without step-by-step solution.
6)~Towers-of-Hanoi-Seq: Solve the three-ring Towers of Hanoi with step-by-step solution.
7)~Packing-Shapes: Place a specified shape in the brown box.
8)~Put-Block-in-Bowl: Place all blocks of a specified color into bowls of a specified color. 
The simulated environments in Task 1-3) employ an xArm6 robot, 
constrained to move in a 2D plane with a cylindrical end-effector.
Task 4-8) utilizes a Universal Robot UR5e with a suction gripper.
More details and success metrics of each task are shown in Appendix A.
}
For each task, we evaluate our method and baselines for 100 episodes with random initial object positions and language instructions.

\textbf{Baselines.}
We compare LLM+A with three baselines:
1)~Naive LLM: Given the same inputs as LLM+A, we directly prompt the LLM to generate decomposed sub-task plans and motion plans without \textit{affordance prompting}. 
2)~ReAct~\cite{yao2023react}: An interactive decision-making approach of LLM by generating both reasoning traces and primitive actions in an interleaved manner. In our implementation, ReAct decides the best action in \{Move Down, Move Up, Move Left, Move Right\} with the robot coordinates and bounding boxes of blocks as inputs. By planning with such low-level actions, we evaluate its capability in motion control. 3)~Code as Policies~\cite{liang2023code}: An LLM-based code generation approach which directly calls pre-defined API to finish the task. In our implementation, after it translates the robotic tasks into programmatic codes, it finally calls one of the basic actions same as ReAct. For both ReAct and Code as Policies, we introduce two example cases in the prompt to help them understand the evaluation tasks.
For all baselines, we use GPT-4 (July) as the base LLM.
All specific prompts of baseline methods are in the Appendix B.2, B.3, and B.4.

\subsection{Results}

\begin{table}
\centering
\caption{Success rates of LLM+A and baselines in pushing tasks from \textbf{Language-Table}.}
\begin{tabular}{ccccc}
    \specialrule{0.08em}{3pt}{5pt}	
    Method & B2P & B2B & Sep & Average  \\    

    \specialrule{0.08em}{3pt}{5pt}	
    Naive LLM & 14\% & 8\%& 70\% & 29\% \\
    \specialrule{0em}{1pt}{1pt}	
    ReAct & 15\% & 3\% & 40\% & 19\% \\  
    \specialrule{0em}{1pt}{1pt}	
    Code as Policies & 22\% & 2\% & 72\% & 32\% \\  
    \specialrule{0em}{1pt}{1pt}	
    LLM+A & \textbf{60\%}  & \textbf{40\%} & \textbf{77\%}  & \textbf{59\%}\\
    \specialrule{0.08em}{3pt}{0pt}
\end{tabular}
\label{performance}
\end{table}

\begin{figure*}[t]
	\centering
	\includegraphics[width=1\linewidth]{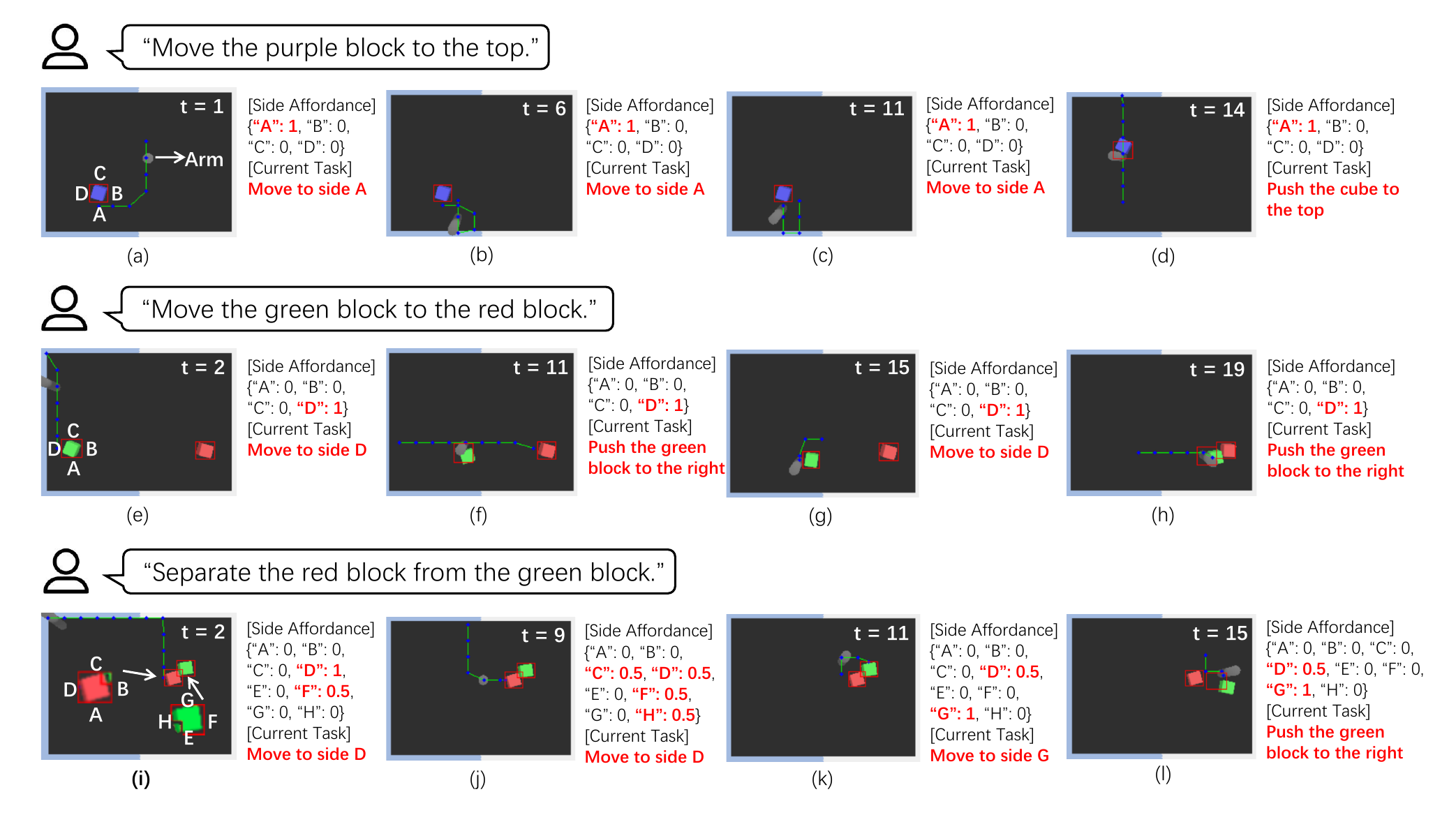}\\
	\caption{Examples of environmental observation and robot trajectories in Block-to-Position (a-d), Block-to-Block (e-h), and Separate (i-l). The gray cylinder indicates the position of the robot end-effector. The blue dots and the green lines represent the waypoints and the planned paths of the control sequences generated by LLM+A, respectively.
 The red boxes denote the detected bounding boxes from Grounding DINO.}\label{tra}
\end{figure*}
\textbf{How effective is LLM+A compared to baselines?}
We present the results of LLM+A and other baselines in three pushing tasks from Langage-Table in Table~\ref{performance}.
\wenzhe{Our method achieves 59\% average success rate across these tasks and outperforms the recent LLM-based approaches significantly, especially in Block-to-Position and Block-to-Block tasks.}
\wenzhe{In particular, we do not provide any demonstration trajectory as in-context examples to prompt LLM. This illustrates our \textit{affordance prompting} approach can effectively simulate the physical interaction inference ability inherent in LLMs, which is essential and valuable for robotics control tasks.}
\wenzhe{A dialogue example of our LLM+A decision process for the Block-to-Position task in Appendix C. When the task is ``push the block to the top'', the robot commanded by our LLM+A approach predicts the bottom side as the correct contacting edge and successfully move it upwards}.
In contrast, baseline methods failed in similar scenrios as shown in Appendix D.
\textit{ReAct} and \textit{Code as Policies} only tend to imitate the procedure of the in-context examples without understanding the physical consequences and make mistakes in choosing the correct side. 
For example, when the task instruction is ``move the block to the upper left corner of the table'', their methods guide the robot to approach the left side of the block and push it towards the left direction. 
% Therefore, although we provide two examples for their prompts, the robot still makes mistakes in choosing the correct side.

\textbf{How well does LLM+A predict affordance values and plan action sequences?}
We present three examples of Block-to-Postion, Block-to-Block, and Separate in Figure~\ref{tra}.
First, we observe that our method can successfully predict goal-conditioned affordance values.
For example, LLM assigns a higher affordance value to the bottom side A of the block to push it to the top side of the table as shown in Figure \ref{tra}(a).
\wenzhe{Besides, the predicted affordance values can dynamically adapt to the changing environment state.}
% Besides, the predicted affordance values also dynamically vary according to the current environment state.
As shown in Figure~\ref{tra}(i), LLM firstly assigns a higher affordance score to the left side D of the red block at the beginning time. 
However, in Figure~\ref{tra}(k), LLM improves the affordance value to the top side G of the green block, since pushing side D could indirectly push the green block off the table.
Second, \wenzhe{our method can consistently decompose the task instruction into easier high-level sub-tasks for better performance.}
% our method can decompose the task instruction and generate the high-level current sub-task for the robot.
As shown in Figure~\ref{tra}(f), the current sub-task is to push the green block to the right when the robot has approached side D.
However, in Figure~\ref{tra}(g), LLM guides the robot to move to side D again when the robot breaks away from the block.
Third, \wenzhe{our method provides a way to translate a high-level instruction into low-level executable control sequences for the robot.}
% our method can generate executable low-level control sequences for the robot.
For example, in Figure~\ref{tra}(c), LLM controls the robot to detour to side A of the purple block to prevent contact with the block in an unexpected direction. Owing to the low-level control sequences being raw coordinates, this proves that our approach is valid to arouse the spatial relationship understanding ability in LLMs, which is essential and promising for further non-training paradigms for robotics control.
More examples of trajectories with different block shapes and interference are shown in Appendix E and F.

\begin{table}
\centering
\caption{Failure case analysis of LLM+A in fours tasks from \textbf{Language-Table}.}
\begin{tabular}{ccccc}
    \specialrule{0.08em}{3pt}{5pt}	
    Failure & B2P & B2B & Sep & Average \\
    \specialrule{0.08em}{3pt}{5pt}	
    Object Detection Fail & 0\% & 6\%  & 6\% & 4\% \\
    \specialrule{0em}{1pt}{1pt}		
    Affordance Prediction Fail & 8\% & 20\%  & 10\% & 13\%\\
    \specialrule{0em}{1pt}{1pt}		
    Task Planning Fail& 8\% & 2\% & 2\% & 4\% \\
    \specialrule{0em}{1pt}{1pt}		
    Motion Controlling Fail& 12\% & 22\%  & 4\% & 13\% \\
    \specialrule{0em}{1pt}{1pt}	
    Exceed Time Budget& 12\% & 18\% & 1\% & 10\% \\    
    \specialrule{0.08em}{3pt}{5pt}	
\end{tabular}
\label{fail_tab}
\end{table}

\wenzhe{We present the failure case analysis of LLM+A in three Language-Table tasks in Table \ref{fail_tab}.
We find that our LLM+A framework seldom make mistakes in high-level task planning. Compared with the low success rate achieved by baseline methods, our design of prompting the LLM to consider possible interaction consequences before planning proves its effectiveness.
Most failures happen in complex affordance prediction and path planning, as in Block-to-Block task. This is because the LLM not only needs to estimate the interaction consequence between the end-effector and the target objects, but also the chain consequences to other objects. Both the chain consequences and fine-grained motion plans require a better knowledge of spatial geometry. As our LLM+A framework is not limited to a specific VLM, we believe it could be further improved with a more powerful multi-modal large models for geometry understanding. Empowering the LLM+A with multi-modal modal for zero-shot robot control would be a potential direction for our future work.}

\begin{table}
\centering
\caption{Success rates of LLM+A in Pick\&Place tasks from \textbf{CLIPORT}. }
\begin{tabular}{cc}
    \specialrule{0.08em}{3pt}{5pt}	
    Task Name & Success Rate \\    

    \specialrule{0.08em}{3pt}{5pt}	
    Stack-Block-Pyramid & 76\%  \\  
    \specialrule{0em}{1pt}{1pt}	
    Towers-of-Hanoi & 80\%  \\
    \specialrule{0em}{1pt}{1pt}	
    Towers-of-Hanoi-Seq & 96\% \\  
    \specialrule{0em}{3pt}{0pt}
    Packing-Shapes & 80\% \\  
    \specialrule{0em}{1pt}{1pt}	
    Put-Block-in-Bowl & 95\%  \\
    \specialrule{0.08em}{1pt}{1pt}	

\end{tabular}
\label{pickplace}
\end{table}

\begin{figure}
	\centering
	\includegraphics[width=0.75\linewidth]{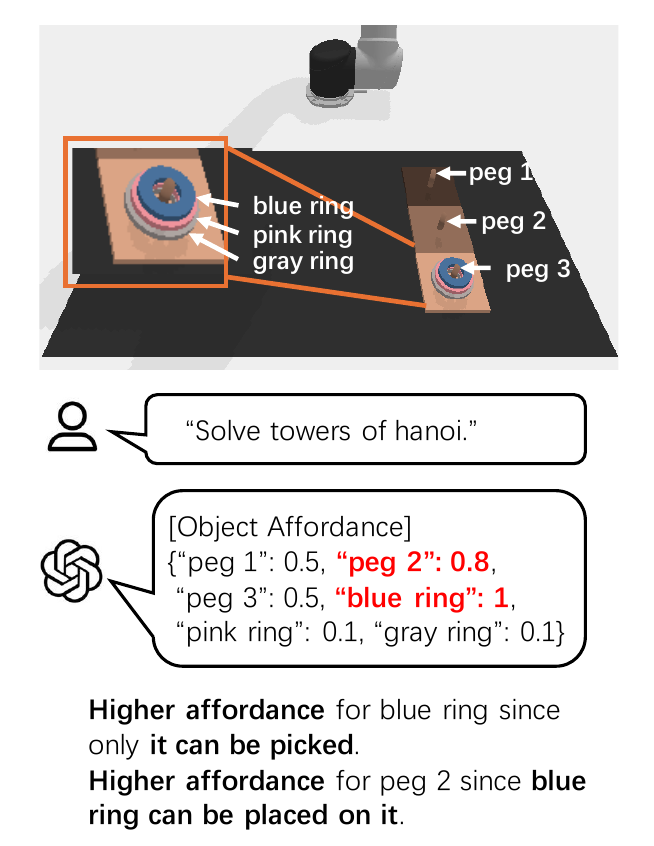}\\
	\caption{Example of environmental observation and affordance prediction from LLM in Towers-of-Hanoi task. }\label{hanoi}
\end{figure}

\textbf{How robust is \textit{affordance prompting} when generalized to heterogeneous tasks?}
\guangran{We transfer LLM+A to five tasks from CLIPORT with completely the same prompts in the pushing tasks to evaluate the robustness of \textit{affordance prompting} on heterogeneous tasks. 
These tasks differ from Language-Table tasks in multiple aspects.
First, the CLIPORT tasks leverages a Universal Robot UR5e different from the xArm6 robot in pushing tasks.
Second, the interactive parts of blocks are block centers for the suction gripper different from multiple sides of blocks in pushing tasks for the cylindrical end-effctor.
Third, the state dynamics of tabletop environments are different in these tasks.
Therefore, LLMs need to adaptively understand the different physical interaction processes according to the robot type and task instructions.
We report the success rates of five tasks in Table \ref{pickplace}.
LLM+A achieves around 85\% average success rates in all scenarios. 
We show an example of environmental observation and affordance prediction from LLM of Towers-of-Hanoi task in Figure \ref{hanoi}.
We can observe that, given the instruction ``solve towers of hanoi'' and the textual environemental observation, LLM assigns higher affordance for the blue ring and the second peg.
The reason is that only the top blue ring can be picked and placed on the second peg, which conforms to the physical laws at the current timestep.
This suggests that \textit{affordance prompting} is a general framework that can help the LLMs understand the attributes and consequences of various robotic tasks, and make it generalizable for dealing with heterogeneous tasks.}
We provide an example dialogue of LLM+A in Appendix C.
The details of prompt design are shown in Appendix B.1.

\section{Conclusion}
In this paper, we propose the LLM+A framework for language-conditioned robotic manipulation tasks with large models, which successfully decomposes the language instruction into several sub-tasks, generates the robot control sequences, and easily extends to heterogeneous tasks. 
This shows the potential of LLMs in simultaneously achieving both planning and motion control, which provides an alternative training-free paradigm for utilizing the LLMs in robotic tasks. 
This significantly mitigates the dataset bottleneck issue for the robotics field. Besides, in order to ground the decomposed sub-tasks in the physical world and guarantee the feasibility of generated control sequences, we prove the proposed \textit{affordance prompting} is crucial to stimulate the physical knowledge from the LLMs about spatial relationship and interaction consequences inference. 
The experimental results demonstrate the effectiveness and generalizability of LLM+A. In the future, we will first optimize the process of using LLM+A to increase the time efficiency and secondly try to apply our method to more broad robotics tasks, including those those tasks involving more complex physical interaction and different robot structures.

\clearpage

\bibliographystyle{named}
\bibliography{ijcai24}

\end{document}